\documentclass[9pt,shortpaper,twoside,web]{ieeecolor}
\usepackage{generic}
\usepackage{cite}
\usepackage{amsmath,amssymb,amsfonts}
\usepackage{algorithmic}
\usepackage{graphicx}
\usepackage{textcomp}
\usepackage{hyperref}
\usepackage{multirow}
\usepackage{booktabs}
\usepackage{bm}
\usepackage{makecell}
\usepackage{hyperref}

\def\BibTeX{{\rm B\kern-.05em{\sc i\kern-.025em b}\kern-.08em
    T\kern-.1667em\lower.7ex\hbox{E}\kern-.125emX}}
\begin{document}
\title{Unsupervised Brain Tumor Segmentation with Image-based Prompts}
\author{Xinru~Zhang, Ni Ou, Chenghao Liu, Zhizheng Zhuo, Yaou Liu, and Chuyang~Ye
\thanks{Xinru Zhang, Chenghao Liu, and Chuyang Ye are with School of Integrated Circuits and Electronics, Beijing Institute of Technology, Beijing, China.}
\thanks{Ni Ou is with School of Automation, Beijing Institute of Technology, Beijing, China.}
\thanks{Zhizheng Zhuo and Yaou Liu are with Beijing Tiantan hospital, Capital Medical University, Beijing, China.}
\thanks{Corresponding authors: Chuyang Ye (chuyang.ye@bit.edu.cn); Yaou Liu (liuyaou@bjtth.org).}
\thanks{Chuyang Ye is supported by the Beijing Municipal Natural Science Foundation (L192058). 
Yaou Liu is supported by the National Natural Science Foundation of China (81870958 \& 81571631), the Beijing Municipal Natural Science Foundation for Distinguished Young Scholars (JQ20035), and the Special Fund of the Pediatric Medical Coordinated Development Center of Beijing Hospitals Authority (XTYB201831).}
}

\maketitle

\begin{abstract}
Automated brain tumor segmentation based on \textit{deep learning}~(DL) has achieved promising performance. 
However, it generally relies on annotated images for model training, which is not always feasible in clinical settings.
Therefore, the development of unsupervised DL-based brain tumor segmentation approaches without expert annotations is desired.
Motivated by the success of \textit{prompt learning}~(PL) in natural language processing, we propose an approach to unsupervised brain tumor segmentation by designing image-based prompts that allow indication of brain tumors, and this approach is dubbed as \textit{PL-based Brain Tumor Segmentation} (PL-BTS). 
Specifically, instead of directly training a model for brain tumor segmentation with a large amount of annotated data, we seek to train a model that can answer the question: is a voxel in the input image associated with tumor-like hyper-/hypo-intensity?
Such a model can be trained by artificially generating tumor-like hyper-/hypo-intensity on images without tumors with hand-crafted designs.
Since the hand-crafted designs may be too simplistic to represent all kinds of real tumors, the trained model may overfit the simplistic hand-crafted task rather than actually answer the question of abnormality.
To address this problem, we propose the use of a validation task, where we generate a different hand-crafted task to monitor overfitting.
In addition, we propose PL-BTS+ that further improves PL-BTS by exploiting unannotated images with brain tumors.
Compared with competing unsupervised methods, the proposed method has achieved marked improvements on both public and in-house datasets, and we have also demonstrated its possible extension to other brain lesion segmentation tasks.
\end{abstract}

\section{Introduction}
\label{sec:intro}
\IEEEPARstart{B}{rain} tumor is one of the most common and deadly diseases of the central nervous system.
For example, \textit{high-grade glioma}~(HGG) is the most common and malignant brain tumor that can grow rapidly, invade surrounding normal tissue, and recur with a median survival time of 18--20 months~\cite{Andrea2022Spatiotemporal,Menze2015Multimodal,Williams2008Science}. 
\textit{Glioblastoma}~(GBM) is the most aggressive form of brain cancer~\cite{Gregory2020Systemic}, currently accounting for 47\% of diagnosed brain cancers~\cite{ALIFIERIS201563}. GBM is characterized by its high invasiveness, poor clinical prognosis, high mortality rate, and frequent recurrence~\cite{Gregory2020Systemic}.

Brain tumor segmentation on \textit{magnetic resonance}~(MR) images allows quantitative evaluation of brain tumors and provides valuable information for better understanding of tumor characteristics and treatment planning~\cite{isensee2021nnu,magadza2021deep,kamnitsas2017efficient}.
Brain tumor segmentation can be achieved with conventional image processing techniques, such as thresholding-based and region-based methods~\cite{gordillo2013state,dubey2009region}, but the accuracy of these simplistic methods is limited.
More recently, \textit{deep learning}~(DL) techniques have been applied to brain tumor segmentation with outstanding segmentation accuracy~\cite{isensee2021nnu,havaei2017brain,wang2021transbts,ding2021rfnet,ding2021tostagan}. The DL-based methods improve the objectiveness, reproducibility, and scalability of quantitative evaluation of brain tumors. 
However, the success of DL-based methods relies on large-scale annotated training data~\cite{magadza2021deep}, which can be costly to collect and reduce the applicability of DL-based brain tumor segmentation in real-world scenarios.
Although approaches based on semi-supervised learning~\cite{cui2019semi,meier2014patient} or self-supervised learning~\cite{zhang2021self,zhuang2019self} could alleviate the need for annotated data, manual annotations are still required for model training.

There are also works that explore DL-based brain tumor segmentation (or brain lesion segmentation that is closely related) without requiring any manual annotation. 
Most of these unsupervised approaches are reconstruction-based methods~\cite{pinaya2022unsupervised,baur2021autoencoders,naval2021implicit,wu2021unsupervised}. In particular, in these methods DL-based models are trained to reconstruct input images drawn from a normal---i.e., healthy---population, and it is assumed that the trained models cannot properly reconstruct images with brain tumors.
Then, for images with brain tumors, the input is compared with the reconstructed version, and anomalous regions are identified based on the reconstruction error.
However, the accuracy of these approaches can still be unsatisfactory, and further exploration of unsupervised brain tumor segmentation without manual annotations is desired.

In this paper, we continue to explore the problem of unsupervised DL-based brain tumor segmentation.
Motivated by the success of \textit{prompt learning} (PL) in zero-shot \textit{natural language processing}~(NLP)~\cite{liu2021pre}, we propose to perform unsupervised brain tumor segmentation by designing image-based prompts that allow indication of brain tumors. 
PL in NLP exploits large pretrained language models~\cite{kenton2019bert,brown2020language} and designs prompt tasks that allow the questions of interest to be addressed by the pretrained models.
The target task is prompted to converge to the task performed by the large pretrained model, and the prompt template on the input data space fits the target task~\cite{bahng2022visual}.
Similar for brain tumor segmentation, the target task should be accomplished with a prompt task that is closely related to it and does not require manual annotations.

Our approach is dubbed as \textit{PL-based Brain Tumor Segmentation} (PL-BTS). PL-BTS comprises two major steps.
First, we design an image-based prompt task that hints the target task of brain tumor segmentation, where given an image with brain tumors a model is trained to answer the question: is a voxel in the input image associated with tumor-like hyper- or hypo-intensity?
To train such a prompt model, we generate artificial hyper-/hypo-intensity with tumor-like shapes on data without brain tumors based on hand-crafted designs. 
Since the hand-crafted designs may be too simplistic to represent all kinds of real tumors, the trained model may overfit the simplistic hand-crafted task~\cite{changpinyo2017predicting} rather than actually answer the question of abnormality.
Because no annotated data is available as the validation set for model selection in the unsupervised setting, addressing the problem of overfitting is challenging.
Therefore, we secondly propose to design a different validation task for model selection, which monitors overfitting.
Compared with the prompt task, the validation task uses a different way of generating artificial hyper-/hypo-intensity.
As during the training process the model tends to first learn generic rules of identifying abnormality~\cite{mutasa2020understanding}, when the accuracy of the validation task begins to decrease, the model is likely to start to overfit the artifact in the data generated with hand-crafted prompt designs.
Thus, model selection is performed based on the turning point of the performance of the validation task to encourage the model to identify generic tumor-like abnormality, and the selected model gives the result of unsupervised brain tumor segmentation.

In addition, we propose PL-BTS+ that extends PL-BTS to the scenario where unlabeled images with brain tumors are available and improves PL-BTS with such unlabeled data. 
Specifically, pseudo-labels are first predicted on the unlabeled images with the prompt model. 
Like the pseudo-labeling strategy that is commonly used in semi-supervised learning~\cite{ghiasi2021simple}, the pseudo-labels can be used to fine-tune the prompt model.
Moreover, as the prompt model tends to undersegment the tumors, the pseudo-labels may contain considerable amounts of false negatives~\cite{li2020analyzing}.
To further improve the fine-tuning of the prompt model, we generate additional synthetic training images by pasting the undersegmented tumor regions represented by the pseudo-labels onto the tumor-free images used for training the prompt model, and this allows removal of the false negatives in the synthetic images.
These synthetic images and the original unlabeled images are used jointly to fine-tune the prompt model together with the pseudo-labels for brain tumor segmentation.

To validate the proposed method, experiments were performed on public and in-house datasets for brain tumor segmentation, where our method was compared with state-of-the-art unsupervised DL-based brain tumor segmentation approaches.
The results show that our method outperforms the competing methods by a large margin in terms of multiple evaluation metrics.
Moreover, using datasets for ischemic stroke lesion segmentation, we demonstrate the potential of generalizing PL-BTS and PL-BTS+ to other brain lesion segmentation tasks.
The code of our method will be released after the work is published.
Our contribution can be summarized as follows:
\begin{itemize}
\item We have proposed PL-BTS for unsupervised brain tumor segmentation that only requires tumor-free images for training. Our method is inspired by prompt learning and provides a perspective that is different from existing reconstruction-based unsupervised methods.
\item In PL-BTS, we have proposed the use of a validation task for model selection, which addresses the problem of having no annotated validation set and effectively avoids overfitting.
\item We have proposed PL-BTS+ that extends PL-BTS to the scenario where unlabeled images with brain tumors are available. In PL-BTS+, an improved pseudo-labeling strategy is developed for better segmentation performance.

\item We have not only validated the proposed method on public and in-house datasets for brain tumor segmentation, but also demonstrated its potential application to other brain lesion segmentation tasks.
In these experiments, the superiority of the proposed method over existing unsupervised methods is shown. 
\end{itemize} 
 
\begin{figure*}[!t]
    \centering
    \includegraphics[width=2.0\columnwidth]{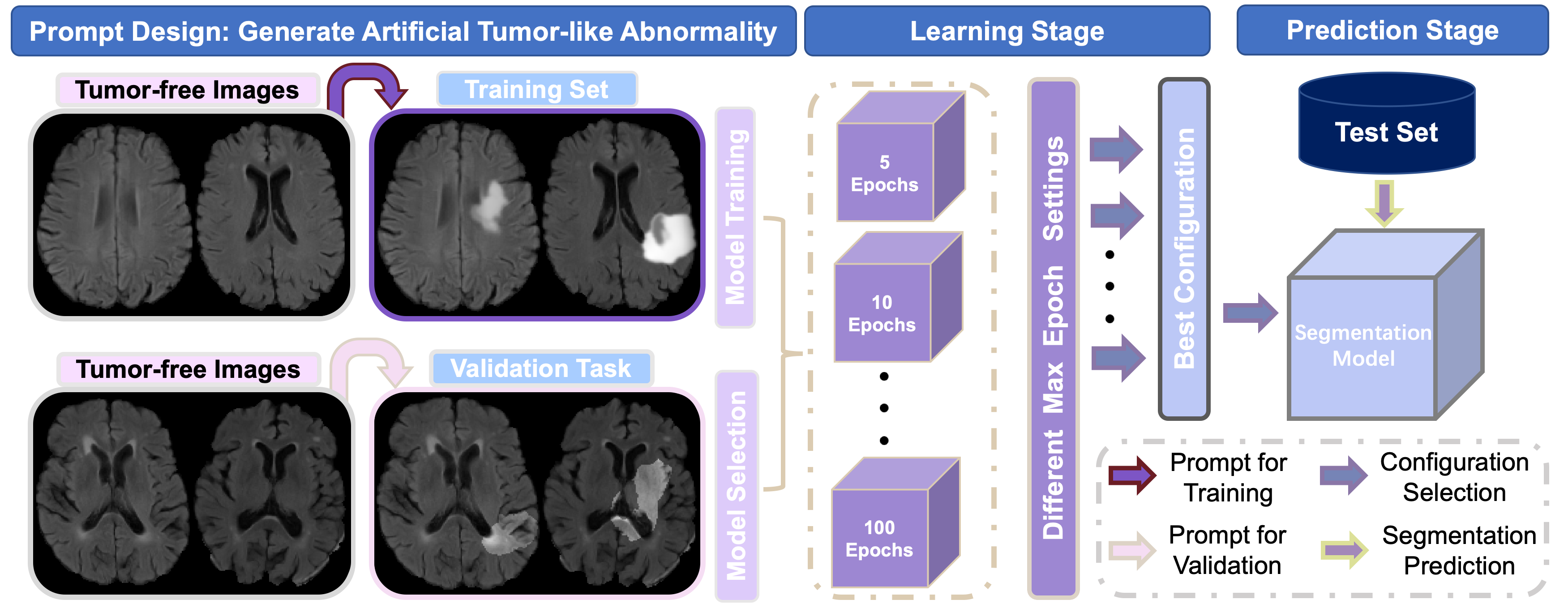}
    \caption{An overview of the PL-BTS method for unsupervised brain tumor segmentation. For the prompt design, we use two different strategies to generate artificial tumor-like hyper-/hypo-intensity on tumor-free images, and the two strategies produce the training set and the data for the validation task respectively. The training set is used to learn the network weights of the prompt model, where multiple maximum numbers of epochs are considered, and the validation task is used for model selection.
    The selected model performs brain tumor segmentation on test data.}
    \label{fig:overview}
\end{figure*}

\section{Related Work}
\subsection{Brain tumor segmentation}
\label{sec:related_tumor}

Brain tumor segmentation can be challenging due to unclear and irregular tumor boundaries caused by partial volume effects and discontinuities~\cite{dong2017automatic}, as well as high variability of tumors \cite{hu2022mutual}. 
Conventionally, simple thresholding-based methods~\cite{gordillo2013state} are developed for the segmentation task, where tumor regions are obtained by comparing their intensities with one or more intensity thresholds; region-based brain tumor segmentation methods~\cite{dubey2009region} are another type of simple methods that rely on region growing~\cite{gordillo2013state}; methods based on voxel classification~\cite{supot2007segmentation,abdel2015brain} use  supervised or unsupervised classifiers to label tumor voxels in the feature space. However, these methods are usually too simplistic to effectively capture the complicated texture of brain tumors and cannot provide satisfactory segmentation quality~\cite{gordillo2013state}.

Due to the success of DL in image processing tasks, DL-based methods~\cite{isensee2021nnu,havaei2017brain,wang2021transbts,ding2021rfnet,ding2021tostagan,jiang2022swinbts} are proposed for brain tumor segmentation and greatly improve the segmentation accuracy. 
In these works, various network architectures are developed based on convolutional neural networks~\cite{isensee2021nnu,havaei2017brain,ding2021rfnet} and/or transformers~\cite{wang2021transbts,jiang2022swinbts,hatamizadeh2022unetr}.
It is also shown that with carefully designed preprocessing, data configuration, and postprocessing, the U-net architecture~\cite{3dunet,2dunet} can already provide high-quality segmentation that is better than or at least comparable to the results achieved with more complex network architectures~\cite{isensee2021nnu}. 

DL-based brain tumor segmentation approaches generally require a large amount of manual annotations for model training, which is infeasible in many clinical and scientific settings. 
To address this problem of data scarcity, semi-supervised and self-supervised methods can be applied, where a large amount of unlabeled data is used together with the scarce annotated data for model training.
For example, semi-supervised methods~\cite{cui2019semi,luo2021dtc} based on the teacher-student framework can be used, where the consistency loss between unlabeled data with different perturbations is added in model training; in self-supervised learning methods~\cite{zhang2021self,chen2019self}, a model is pretrained on unlabeled data with hand-crafted pretext tasks that do not require manual annotations, and it is then fine-tuned with the limited labeled data. 
However, these methods do not fully remove the requirement for manual annotations.

Unsupervised DL-based approaches have been explored for annotation-free brain tumor segmentation~\cite{pinaya2022unsupervised,baur2021autoencoders,naval2021implicit,wu2021unsupervised,dey2021asc}, where no manual annotation is available for model training.
The most popular strategy adopted by these methods is the reconstruction-based strategy, where a model is trained to reconstruct images of healthy subjects and the voxel-level residual between an image with the brain tumor and its reconstructed counterpart is used to indicate the brain tumor. 
The reconstruction models include the \textit{autoencoder}~(AE)~\cite{baur2021autoencoders}, \textit{variational autoencoder}~(VAE)~\cite{baur2021autoencoders}, \textit{vector quantized variational autoencoder}~(VQ-VAE)~\cite{marimont2021anomaly,wang2020image}, generative adversarial networks~\cite{wu2021unsupervised,schlegl2019fAnogan}, and transformer-based networks~\cite{pinaya2022unsupervised,git2022Ahmed}.
In these reconstruction-based unsupervised methods, post-processing of the residual image is required for the final segmentation.
However, the design of effective post-processing is nontrivial, and the performance of the unsupervised brain tumor segmentation methods can still be unsatisfactory.

\subsection{Prompt learning}


Prompt learning has emerged as a promising paradigm for zero-shot learning in NLP. A typical example is PL-based zero-shot emotion analysis.
A prompt template can be built as ``It was [MASK]'' after a raw input sentence such as "I love this movie." 
Then, a large pretrained mask language model, such as BERT~\cite{kenton2019bert} and GPT-3~\cite{brown2020language}, can be applied to fill the blank for emotion analysis directly without retraining or fine-tuning on the target dataset again. 

The application of PL to zero-shot image classification or segmentation has been explored as well so that no manual annotation is required for the task of interest~\cite{bahng2022visual,xu2021simple,ding2022decoupling,wang2022cris}.
In these methods, a vision-language model~\cite{radford2021CLIP} is trained with a huge amount of paired image and text data, which aligns the features of the two modalities.
For zero-shot image classification, the vision-language model associates the image with the most relevant textual representation and produces the classification result based on the text. 
The idea is further extended to semantic segmentation~\cite{xu2021simple,wang2022cris}.
In~\cite{xu2021simple}, an image is first partitioned into different regions, and each region is fed into the vision-language model for classification, which gives the segmentation result without requiring manual annotations for training.
In~\cite{wang2022cris}, a vision-language decoder is further added to the vision-language model to relate fine-grained semantic information from textual representations to each pixel-level activation, which allows zero-shot image segmentation.

The application of vision-language models to medical imaging is still largely vacant, probably due to the lack of pretrained vision-language models designated for medical applications.
It may also be possible to design image-based prompts without requiring language for medical imaging, including brain tumor segmentation, but this direction of research has not been explored yet.

\section{Methods}
\label{sec:Method}

An overview of the proposed PL-BTS method is shown in Fig.~\ref{fig:overview}, where a prompt model that indicates the existence of tumor-like abnormality is trained for unsupervised brain tumor segmentation without requiring any annotated data.
We first introduce the design of the image-based prompt and then describe the proposed use of a validation task for model selection, which avoids overfitting to the prompt task.
Besides the workflow in Fig.~\ref{fig:overview}, we present PL-BTS+ that extends PL-BTS to the scenario where unlabeled data with brain tumors is available and incorporated in model training.

\subsection{PL-BTS for unsupervised brain tumor segmentation}
\label{sec:pl}


\subsubsection{Prompt design}
\label{sec:prompt for training}

Since brain tumors generally appear as hyper- or hypo-intensity on MR images, in PL-BTS the image-based prompt is designed to indicate whether there is hyper-/hypo-intensity at a voxel.
Equivalently, a prompt model is trained to determine for each voxel of an input image: does the voxel in the input image have hyper-/hypo-intensity?
To train such a model, we generate artificial hyper-/hypo-intensity on images without brain tumors with hand-crafted rules, where no manual annotation is needed.

Suppose we are given a set $\mathcal{X}=\{\mathbf{X}_{i}\}_{i=1}^{N}$ of images without brain tumors, where $\mathbf{X}_{i}$ is the $i$-th image and $N$ is the total number of these images.
Based on $\mathbf{X}_{i}$, we can generate a synthetic image $\mathbf{X}$ with regions of hyper-/hypo-intensity and obtain the corresponding mask $\mathbf{Y}$ of abnormality indicating the hyper-/hypo-intensity as
\begin{equation}
\begin{aligned}
\label{eq:prompt_imgs}
\mathbf{X} &= T(\mathbf{X}_{i})\odot \mathbf{A} +  \mathbf{X}_{i}\odot (1-\mathbf{A}), \\
\mathbf{Y}&=  \begin{cases}

1,&\ \mathbf{A} \geq a  \\ 
0,& \ \mbox{otherwise}

\end{cases}.
\end{aligned}
\end{equation}
Here, $T(\cdot)$ represents an intensity transformation that generates hyper-/hypo-intensity, $\mathbf{A}$ is a weight image with intensities ranging from zero to one, $\odot$ represents voxelwise multiplication, and $a$ is a threshold that determines the mask $\bf{Y}$ based on $\bf{A}$.
Eq.~(\ref{eq:prompt_imgs}) mixes the original image without brain tumors and the artificial hyper-/hypo-intensity to generate an image with intensity abnormality, and the threshold $a$ (empirically set to 0.1) is used to account for the partial volume effect in the buffering region where normal tissue and abnormality are mixed.
The key to the prompt design is the designs of $T(\cdot)$ and $\mathbf{A}$, which determine the appearance and shape of the hyper-/hypo-intensity, respectively. 
Note that we do not seek to generate realistic tumors on the images, as it is extremely challenging, if possible at all, without any annotation.

As brain tumors comprise large clusters of abnormally bright signals or dark signals, $T(\cdot)$ is designed to generate hyper-/hypo-intensity with smooth textures. 
Specifically, $T(\cdot)$ blurs and scales the image to produce hyper-/hypo-intensity as 
\begin{eqnarray}
\label{eq:tumor_trans}
T(\mathbf{X}_{i}) &=& \lambda \cdot \mathbf{Blur}(\mathbf{X}_{i}).
\end{eqnarray}
Here, $\mathbf{Blur}(\mathbf{X}_{i})$ is a 3D Gaussian blurring function applied to $\mathbf{X}_{i}$, and $\lambda$ controls the magnitude of the hyper-/hypo-intensity.
The specification of $\lambda$ is introduced later.
Since $T(\mathbf{X}_{i})\odot \mathbf{A}$ is added to $\mathbf{X}_{i}\odot (1-\mathbf{A})$, hyper- or hypo-intensity is created in $\mathbf{X}$.


The mixing weight image $\bf{A}$ that determines the region of hyper-/hypo-intensity is designed to comprise both a tumor-like region and a buffer layer that represents the partial volume effect between normal tissue and regions with hyper-/hypo-intensity.
To this end, we first generate a polyhedron with variable orientation and shape following \cite{perez2021self}. 
The size of the polyhedron is sampled from the uniform distribution $U(20000~\mbox{mm}^3,80000~\mbox{mm}^3)$~\cite{zhang2021self,perez2021self}. 
The center of the polyhedron is randomly placed within the brain, and the intersection of the inserted polyhedron and the brain mask is denoted by $\mathbf{M}$, which also represents the region with hyper-/hypo-intensity. 
To take the partial volume effect into consideration, 3D Gaussian filtering with a kernel size of 1 mm is applied to $\mathbf{M}$, and the filtered result $\bf{A}$ is used for the image generation in Eq.~(\ref{eq:prompt_imgs}).



Brain tumors are generally segmented on T2-weighted or FLAIR images~\cite{Menze2015Multimodal}, where they appear as a cluster of hyper-intensity or also with a relatively dark component in the cluster.
To generate images with tumor-like hyper-intensity, we apply Eq.~(\ref{eq:prompt_imgs}) with 
$\lambda>1$ sampled from the uniform distribution $U(1.5, 5)$.
To further obtain images with a dark component, after applying Eq.~(\ref{eq:prompt_imgs}), we use the following equations for image generation
\begin{eqnarray}
\begin{aligned}
\label{eq:prompt_subregion}
\widetilde{\mathbf{X}} &= \widetilde{T}(\mathbf{X}_{i})\odot \widetilde{\mathbf{A}} +  \mathbf{X}\odot (1-\widetilde{\mathbf{A}}), \\
\widetilde{\mathbf{Y}}&= \mathbf{Y}.
\end{aligned}
\end{eqnarray}
Here, $\widetilde{T}(\mathbf{X}_{i})$ is a different transformation $\widetilde{T}(\mathbf{X}_{i}) = \tilde{\lambda} \cdot \mathbf{Blur}(\mathbf{X}_{i})$, where $\tilde{\lambda}$ is sampled from $U(0.8, 1.2)$ to create the dark region (darker than the hyper-intensity); $\widetilde{\mathbf{A}}$ is another weight image that represents the dark component and is obtained like $\mathbf{A}$\footnote{The polyhedron associated with $\widetilde{\mathbf{A}}$ is constrained to be smaller than that associated with $\mathbf{A}$. Its center is randomly placed within $\mathbf{M}$, and its intersection with $\mathbf{M}$ is smoothed to obtain $\widetilde{\mathbf{A}}$.};
$\widetilde{\mathbf{X}}$ and $\widetilde{\mathbf{Y}}$ are the synthetic image and the corresponding mask indicating abnormality, respectively. 
In this way, when Eqs.~(\ref{eq:prompt_imgs}) and~(\ref{eq:prompt_subregion}) are sequentially applied, tumor-like hyper-intensity with a dark component is generated.


By repeating the image generation procedure in Eq.~(\ref{eq:prompt_imgs}) or using the combination of Eqs.~(\ref{eq:prompt_imgs}) and (\ref{eq:prompt_subregion}) with randomly drawn images, a number of synthetic images with approximately tumor-shaped abnormal intensity can be generated
(see the examples in Fig.~\ref{fig:overview} associated with ``Training Set''). 
The probability of applying Eq.~(\ref{eq:prompt_imgs}) alone or both Eqs.~(\ref{eq:prompt_imgs}) and (\ref{eq:prompt_subregion}) is set to 0.9 or 0.1, respectively.
These synthetic images and the corresponding masks of abnormality are used to train the prompt model.

\subsubsection{Model selection with a validation task}
\label{sec:validation-task}

The abnormality generated with the procedures above can be too simplistic compared with real tumors.
Thus, the prompt model obtained after training convergence can overfit the simplistic synthetic abnormality and generalize poorly to indicate real brain tumors.
Ideally, to avoid overfitting, a validation set with annotated data is used to monitor the training process~\cite{isensee2021nnu}. 
However, in the unsupervised setting there is no annotated data as the validation set.

To address this problem, we assume that during model training, the model first learns generic rules that apply to both simplistic synthetic abnormality and real tumors and then starts to overfit the detailed texture of synthetic abnormality~\cite{changpinyo2017predicting}.
Based on the assumption, we propose to use a validation task to prevent overfitting.
This validation task also aims to identify abnormal intensity generated with hand-crafted rules, but the design of the rules is different from the prompt task for training, so that the two tasks require similar generic knowledge but the task details are different.
In this way, after the generic knowledge about identifying abnormal intensity is sufficiently learned and the prompt model starts to overfit, the performance of the validation task will degrade, and based on this model selection can be performed.

\begin{figure}[!t]
    \centering
    \includegraphics[width=1.0\columnwidth]{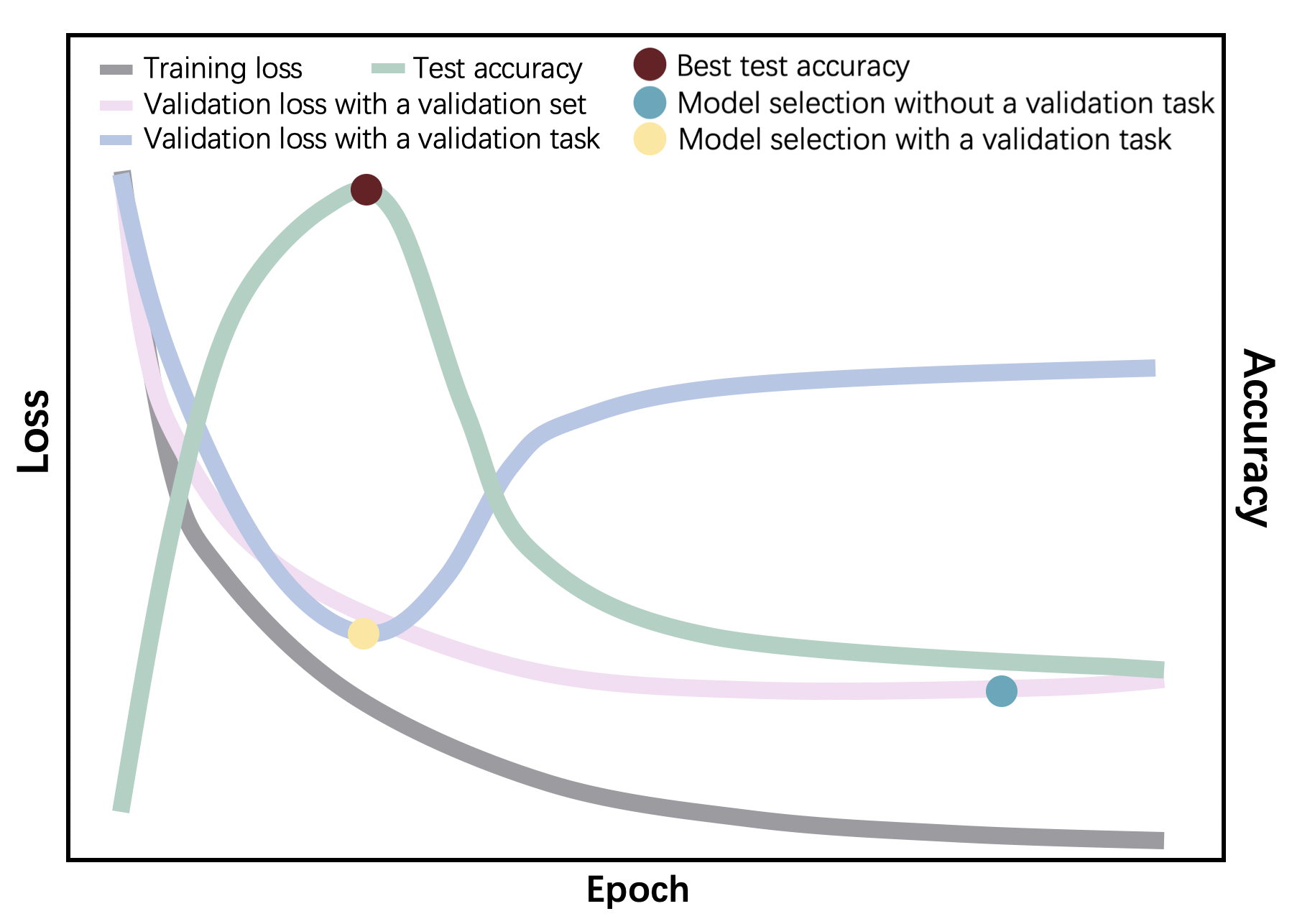}
    \caption{A schematic of the training process of the prompt model. Without the validation task, model selection occurs after the turning point of the test accuracy and is associated with overfitting, even if a validation set comprising the synthetic data is split from the training set for model selection; with the validation task, the turning point can be better identified for model selection.}
    \label{fig:loss}
\end{figure}

A schematic of the benefit of the validation task is shown in Fig.~\ref{fig:loss}.
In the beginning epochs, the prompt model learns useful knowledge from the synthetic data, where both the training loss decreases and the accuracy of brain tumor segmentation on test data increases.
Later, the model starts to overfit, where the training loss still decreases but the test accuracy starts to drop rapidly.
Without the validation task, model selection occurs after the turning point of the test accuracy and is associated with overfitting, even if a validation set comprising the synthetic data is split from the training set for model selection; with the validation task, the turning point can be better identified for model selection.

The detailed design of the validation task is described as follows, which is similar to yet different from the prompt task.
Specifically, we replace the transformation $T(\mathbf{X}_{i})$ and weight image $\mathbf{A}$ in Eq.~(\ref{eq:prompt_imgs}) with
\begin{eqnarray}
\label{eq:validation task1}
{T}(\mathbf{X}_{i}) = \lambda \cdot \mathbf{X}_{i} \mbox{\quad and \quad}
\mathbf{A}= \mathbf{M}.
\end{eqnarray}
Compared with the prompt task, in the validation task no blurring is performed for image generation, and thus the validation task is similar to the prompt task but has different image textures near and in the tumors.
Note that the set $\mathcal{X}$ is split into two subsets for the prompt task and validation task.

Since overfitting is more likely to occur when more training epochs are used and the scheduling of the optimization algorithm, such as Adam~\cite{Diederik2015Adam}, is dependent on the maximum number $T$ of epochs, we considered different choices of $T$ for training the prompt model.
Specifically, the set of candidate values of $T$ is $\{100,50,20,10,5\}$, and $T$ is also selected based on the performance of the validation task.


\subsection{PL-BTS+: further improvement by incorporating unlabeled images with brain tumors}
\label{sec:unlabeled}

In practice, a large amount of unlabeled images with brain tumors may be available, and they can be helpful in semi-supervised settings~\cite{cui2019semi}, where both labeled and unlabeled images are used for model training.
Motivated by this, we hypothesize that these unlabeled images can also benefit unsupervised brain tumor segmentation.
For example, the pseudo-labeling strategy~\cite{ghiasi2021simple,zoph2020rethinking} in semi-supervised learning can be adapted for our purpose, where the unlabeled images and the pseudo-labels predicted by the prompt model on them are used together for model training. 
However, as unsupervised segmentation is challenging due to the complete lack of annotated training data, the pseudo-labels may undersegment the tumor and contain many false negatives. 

\begin{figure}[!t]
    \centering
    \includegraphics[width=1.0\columnwidth]{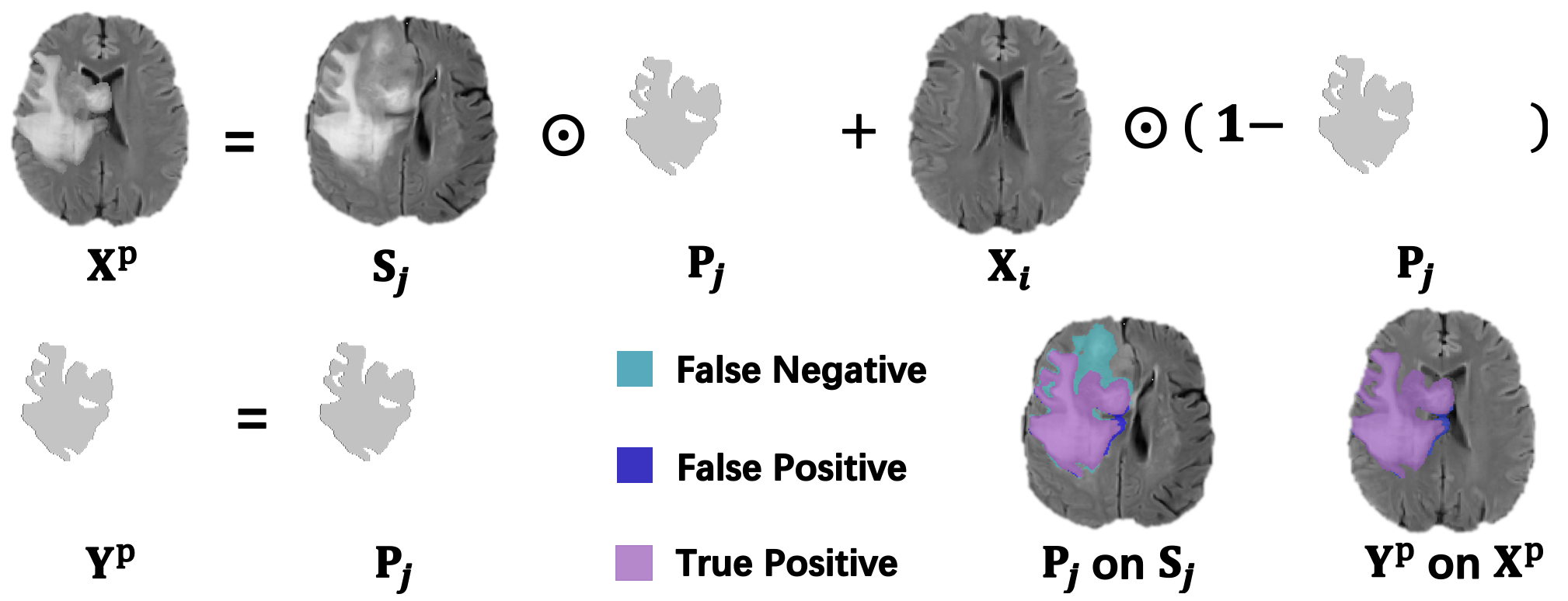}
    \caption{A graphical illustration of the pasting of the undersegmented tumors represented by the pseudo-labels onto the images without brain tumors. The pseudo-label of the original unlabeled image can comprise many false negative voxels due to undersegmentation.
    After the pasting the false negatives are reduced and the quality of the pseudo-label is improved.}
    \label{fig:underseg}
\end{figure}

To alleviate this problem, we propose PL-BTS+, where the undersegmented tumors represented by the pseudo-labels are pasted onto the images without tumors that are used to train the prompt model.
In this way, the false negatives can be reduced in the resulting synthetic images. 
Mathematically, suppose we are given a set $\mathcal{S}=\{\mathbf{S}_{j}\}_{j=1}^{M}$ of $M$ unlabeled images with brain tumors, where $\mathbf{S}_{j}$ is the $j$-th unlabeled image; then the pasting operation is represented as
\begin{eqnarray}
\label{eq:carveX}
{\mathbf{X}}^{\mathrm{p}} &=& \mathbf{S}_{j}\odot \mathbf{P}_{j} + \mathbf{X}_{i}\odot (1 - \mathbf{P}_{j}), \\
{\mathbf{Y}}^{\mathrm{p}} &=& \mathbf{P}_{j},
\label{eq:carveY}
\end{eqnarray}
where $\mathbf{X}^{\mathrm{p}}$ and $\mathbf{Y}^{\mathrm{p}}$ are the image with a pasted brain tumor and the corresponding tumor pseudo-label, respectively, $\mathbf{P}_{j}$ is the pseudo-label of $\mathbf{S}_{j}$ predicted by the prompt model, and $\mathbf{X}_{i}$ is the $i$-th image without tumors defined in Section~\ref{sec:pl}.
A graphical illustration of the pasting is shown in Fig.~\ref{fig:underseg}.
Note that here we assume that $\mathbf{X}_{i}$ and $\mathbf{S}_{j}$ have the same dimension, and if it is not the case we can simply resize $\mathbf{S}_{j}$ and $\mathbf{P}_{j}$ to apply Eq.~(\ref{eq:carveX}).

A number of images $\mathbf{X}^{\mathrm{p}}$ and pseudo-labels $\mathbf{Y}^{\mathrm{p}}$ can be generated by repeating Eqs.~(\ref{eq:carveX}) and (\ref{eq:carveY}) with randomly drawn $\mathbf{S}_{j}$ and $\mathbf{X}_{i}$.
These images and pseudo-labels are used together with the original unlabeled images and their pseudo-labels to fine-tune the prompt model for brain tumor segmentation.
The fine-tuning allows the model to further learn from real tumor textures.
Since the prompt model has already been well trained, only ten epochs are used for fine-tuning.

\subsection{Details about the segmentation framework}

Our method is agnostic to the architecture of the segmentation network and the segmentation framework.
For demonstration, we select nnU-Net~\cite{isensee2021nnu} that has achieved state-of-the-art performance across a variety of medical image segmentation tasks as our segmentation framework.
nnU-net uses the U-net architecture~\cite{3dunet,2dunet} and automatically determines the data configuration and hyperparameter settings.
For more details about nnU-net, we refer readers to~\cite{isensee2021nnu}.

\section{Results}
\label{sec:Experiments}

To evaluate the proposed method, we performed experiments on two datasets for brain tumor segmentation. In addition, to show that PL-BTS and PL-BTS+ can be extended beyond brain tumor segmentation, we also considered a different brain lesion segmentation task, ischemic stroke lesion segmentation, for evaluation. We first describe the datasets considered in the experiments and the experimental settings. Then, we introduce the competing unsupervised methods for comparison.
Finally, the detailed evaluation results are presented, including the comparison of the proposed method with competing methods and the investigation of the benefit of incorporating unlabeled images for training.

\subsection{Data description and experimental settings}
\label{sec:data}

For brain tumor segmentation, we used the FLAIR images from the public dataset OASIS~\cite{marcus2010open} as the images without brain tumors for training the prompt model, and evaluated the performance of the proposed method on the public dataset BraTS2021~\cite{baid2021rsna} and an in-house dataset BTFLAIR.
For the extension to ischemic stroke lesion segmentation, we used the \textit{diffusion weighted images}~(DWIs) from an in-house dataset BTDWI and the public dataset ISLES2022~\cite{ISLES} as the images for training the prompt model, and evaluated the proposed method on ISLES2022.
The acquisition of the in-house datasets was approved by the institutional review board of Beijing Tiantan Hospital, Capital Medical University, with written informed consent obtained from each participant according to the Declaration of Helsinki.
The details about these public and in-house datasets and their experimental settings are given below.

\subsubsection{OASIS}
The OASIS dataset~\cite{marcus2010open} is a publicly available dataset for neuroimaging study and analysis.
There are no images with brain tumors in the dataset, which meets our need for prompt learning. 
We randomly selected 100 FLAIR images from the dataset and generated one image with abnormality from each of them (100 in total) for the prompt task; 100 different FLAIR images were also selected from the dataset, and one image with abnormality was generated from each of them (100 in total) for the validation task. 
The FLAIR images have been registered to the corresponding T1-weighted images and skull-stripped with BET~\cite{smith2002fast}. 
All these images have the same voxel size of 1 mm isotropic.

\subsubsection{BraTS2021}

The BraTS2021 dataset~\cite{baid2021rsna} comprises multimodal MR images with brain tumors acquired from multiple institutions under standard clinical conditions.
There are 1251 cases of which both the images and tumor annotations are publicly available. 
The multimodal images of the same subject have been co-registered and skull-stripped.
All these images have the same voxel size of 1 mm isotropic.
In our experiment, the FLAIR images were selected for the segmentation of the whole tumor.
All 1251 cases were used as the test set to evaluate the performance of the prompt model, as no annotation was required to train the prompt model.

As described in Section~\ref{sec:unlabeled}, our method allows the incorporation of unlabeled images with brain tumors into the training process. To evaluate the benefit of the incorporation, 100 images were randomly selected from the 1251 cases and included in training without using their annotations. 
The pseudo-labels predicted by the prompt model on the 100 images were pasted onto the 200 FLAIR images from the OASIS dataset.
Note that the combination of the pseudo-label and the OASIS image was randomly selected.
We ensured that each of the 200 FLAIR images was used once and each pseudo-label was used twice.
This led to 200 synthetic images with pseudo-labels for model fine-tuning.
In this case, the remaining 1151 cases were used as the test set.

\subsubsection{BTFLAIR}
BTFLAIR is an in-house dataset provided by Beijing Tiantan Hospital for segmenting whole brain tumors (ethical approval reference number: KY2022-078-04). 
The dataset includes 67 annotated FLAIR images acquired on multiple scanners, and they have been skull-stripped with  BET~\cite{smith2002fast}. 
The voxel size of these images ranges from $0.875 \mbox{ mm} \times 0.875 \mbox{ mm}  \times 2 \mbox{ mm}$ to $2 \mbox{ mm} \times 2 \mbox{ mm}  \times 5 \mbox{ mm}$.
Like the BraTS2021 dataset, all images in BTFLAIR were used as the test set to evaluate the performance of the prompt model.

\subsubsection{BTDWI}
\label{sec:data_btdwi}

BTDWI is an in-house DWI dataset provided by Beijing Tiantan Hospital (ethical approval reference number: 2021-KY-0112). 
There are 47 DWIs acquired with a $b$-value of 1000 s/$\mbox{mm}^2$, and the skull was removed for each DWI with BET~\cite{smith2002fast}.
The voxel size of the DWIs is around $0.5 \mbox{ mm} \times 0.5 \mbox{ mm}  \times 2 \mbox{ mm}$.
There are no ischemic stroke lesions on the DWIs, and the 47 images were combined with the three images without ischemic stroke lesions described next in Section~\ref{sec:data_isles} (50 in total) to train the prompt model that segments ischemic stroke lesions.
25 of the 50 DWIs without ischemic stroke lesions were used for the prompt task, and the other 25 DWIs were used for the validation task.
We generated four images with abnormality from each DWI.
Note that as the size of ischemic stroke lesions is different from that of brain tumors~\cite{hakimelahi2014time}, during image generation the size of the polyhedron was adjusted to be sampled from a mixture of two uniform distributions $\frac{1}{2}U(200~\mbox{mm}^3,300~\mbox{mm}^3)+\frac{1}{2}U(5000~\mbox{mm}^3,30000~\mbox{mm}^3)$; also, as ischemic stroke lesions appear as hyper-intensity on DWIs, no dark component was used for image generation.

\subsubsection{ISLES2022}
\label{sec:data_isles}
The ISLES2022 dataset~\cite{ISLES} is publicly available and comprises 250 cases of ischemic stroke patients. 
We used the skull-stripped DWIs in the dataset, which were acquired on different scanners with a $b$-value of 1000 s/$\mbox{mm}^2$.
The voxel size of these images ranges from $0.23 \mbox{ mm} \times 0.23 \mbox{ mm}  \times 6 \mbox{ mm}$ to $0.86 \mbox{ mm} \times 0.86 \mbox{ mm}  \times 7 \mbox{ mm}$.
There are three cases that contain no ischemic stroke lesions.
They were used for training the prompt model together with the BTDWI dataset as described in Section~\ref{sec:data_btdwi}.
The other 247 DWIs that contain ischemic stroke lesions were all used as the test set to evaluate the performance of the prompt model.

\begin{figure*}[!t]
    \centering
    \includegraphics[width=1.8\columnwidth]{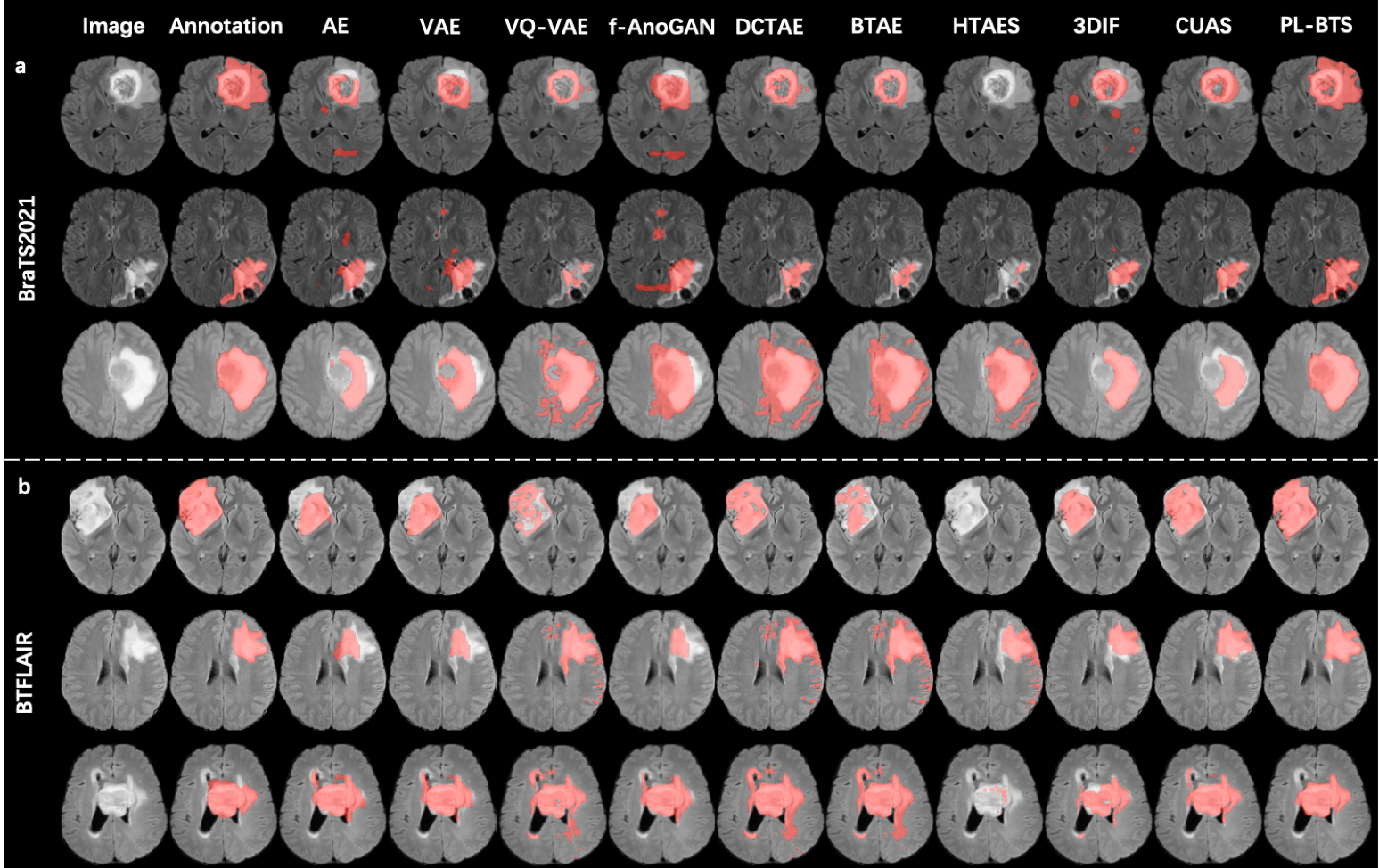}
    \caption{Examples of brain tumor segmentation results (red) on test scans for (a) BraTS2021 and (b) BTFLAIR. The results of PL-BTS and each competing method are overlaid on the FLAIR image. The expert annotation is also shown for reference.}
    \label{fig:BraTS021&BTFLAIR}
\end{figure*}

Again, to further investigate the benefit of incorporating unlabeled images with ischemic stroke lesions in training, we randomly selected 100 DWIs from the 247 DWIs with ischemic stroke lesions and added them into the training process as described in Section~\ref{sec:unlabeled} without using their annotations.
The pseudo-label predicted by the prompt model on each of the 100 DWIs was randomly pasted onto the 50 DWIs without ischemic stroke lesions from the BTDWI and ISLES2022 datasets.
We ensured that each pseudo-label was used twice and each DWI without ischemic stroke lesions was used four times.
This resulted in 200 synthetic images with ischemic stroke lesions for model fine-tuning.
In this case, the remaining 147 DWIs were used as the test set. 
\subsection{Competing methods}

Our method was compared with nine existing unsupervised DL-based brain anomaly detection methods that can be used to segment brain tumors or ischemic stroke lesions.
These methods include typical autoencoder-based approaches~\cite{baur2021autoencoders} (AE, 
VAE, and VQ-VAE), a generative adversarial network f-AnoGAN~\cite{baur2021autoencoders}, more recent methods based on Transformer~\cite{git2022Ahmed} (\textit{dense convolutional transformer autoencoder}~(DCTAE), \textit{basic vision transformer autoencoder}~(BTAE), and \textit{hierarchical transformer autoencoder with skip connections}~(HTAES)), and two other approaches \textit{3D implicit fields}~(3DIF)~\cite{naval2021implicit} and \textit{constrained unsupervised anomaly segmentation}~(CUAS)~\cite{silva2022constrained}  that are designed for unsupervised brain tumor segmentation.

The images without brain tumors or ischemic stroke lesions specified in Section~\ref{sec:data} were used by the competing methods for model training.
Note that in our experiments, we focused on the strictly unsupervised setting, where no validation set with manual annotations was available, and this setting was also applied to the competing methods.

All the competing methods are open-source\footnote{AE, VAE, and f-AnoGAN can be obtained from \url{https://github.com/StefanDenn3r/Unsupervised_Anomaly_Detection_Brain_MRI}; VQ-VAE, BTAE, DCTAE, and HTAES are available at  \url{https://github.com/ahmedgh970/Transformers_Unsupervised_Anomaly_Segmentation}; 3DIF is available at \url{https://github.com/snavalm/ifl_unsup_anom_det}; CUAS is available at \url{https://github.com/jusiro/constrained_anomaly_segmentation}.}.
We used their publicly available code and default settings for our experiments, except that the maximum number of training epochs was set to 100 like in the proposed method, and this number was found to be sufficient for the competing methods.

\begin{table*}[!ht]
\renewcommand\arraystretch{1.05}
\centering
\caption{Means and standard deviations of the Dice coefficient (\%), voxelwise precision (\%), and voxelwise recall (\%) of the segmentation results of PL-BTS and the competing methods on the test set for BraTS2021 and BTFLAIR. Asterisks indicate that the difference between PL-BTS and the competing method is statistically significant (***: $ p< 0.001$) using a paired Student's $t$-test. The performance of PL-BTS achieved without using the validation task (PL-BTS w/o VT) is also shown. The best results are highlighted in bold.}
\label{tab:OASIS}
\resizebox{0.9\textwidth}{!}{
\begin{tabular}{ccccccc}

\toprule
                          & \multicolumn{3}{c}{\textbf{BraTS2021 (1251 cases)}} & \multicolumn{3}{c}{\textbf{BTFLAIR (67 cases)}}                                                        \\ \cmidrule(lr){2-7}
\multirow{-2}{*}{Method} & Dice      & Precision     & Recall   & Dice      & Precision     & Recall \\
\midrule
 
AE      & 33.64 $\pm$ 21.45*** & 62.09 $\pm$ 30.46*** & 30.00 $\pm$ 19.37*** & 41.80 $\pm$ 20.62***                 & 57.92 $\pm$ 29.93***                      & 38.90 $\pm$ 21.20***                   \\
VAE     & 42.67 $\pm$ 18.30***     & 56.53 $\pm$ 25.18***         & 39.63 $\pm$ 19.66***      & 42.26 $\pm$ 21.12***                       & 59.24 $\pm$ 32.12***                            & 39.86 $\pm$ 20.98***                         \\
VQ-VAE      & 31.62 $\pm$ 26.24***     & 74.38 $\pm$ 34.02***         & 30.98 $\pm$ 29.73***      & 30.41 $\pm$ 25.24***                      & 51.28 $\pm$ 39.34***                            & 45.41 $\pm$ 33.33***                         \\
f-AnoGAN    & 39.36 $\pm$ 19.00***     & 36.65 $\pm$ 22.94***         & 53.51 $\pm$ 18.48***      & 33.37 $\pm$ 24.19***                       & 47.34 $\pm$ 37.79***                           & 40.29 $\pm$ 26.43***                        \\
DCTAE       & 38.65 $\pm$ 29.06***     & 74.40 $\pm$ 34.11***         & 40.56 $\pm$ 34.46***      & 34.67 $\pm$ 28.94***                       & 50.40 $\pm$ 39.87***                            & 54.87 $\pm$ 35.99***                         \\
BTAE        & 30.54 $\pm$ 30.92***     & 64.04 $\pm$ 41.71***         & 32.86 $\pm$ 35.76***    & 28.46 $\pm$ 28.21***                       & 52.66 $\pm$ 41.92***                           & 47.97 $\pm$  38.87***                        \\
HTAES                     & 14.57 $\pm$ 24.07***           & 35.36 $\pm$ 44.04***              &15.37 $\pm$ 27.48*** &08.44 $\pm$ 17.29***        &25.00 $\pm$ 20.35***        & 17.81 $\pm$ 32.34***                        \\
3DIF    & 47.10 $\pm$ 17.21***     & 54.51 $\pm$ 24.54***         & 48.79 $\pm$ 16.52***      & 32.07 $\pm$ 17.52***                       & 49.15 $\pm$ 32.01***                            & 36.47 $\pm$ 23.93***                         \\
CUAS        & 54.74 $\pm$ 18.57***           & 74.16 $\pm$ 25.12***              &47.71 $\pm$ 19.23*** &44.19 $\pm$ 28.14***        &59.69 $\pm$ 35.45***        & 46.15 $\pm$ 28.89***                        \\

PL-BTS                    & \textbf{79.56 $\pm$  15.08}          & \textbf{80.36 $\pm$ 15.90}              &\textbf{82.90 $\pm$ 17.77}            &\textbf{70.60 $\pm$ 20.14}                             & {72.91 $\pm$ 23.38}                                 &\textbf{77.42 $\pm$ 22.29}    
\\
\hline
PL-BTS w/o VT                   & {40.94 $\pm$ 30.33}          & {71.70 $\pm$ 34.83}              &{32.70 $\pm$ 28.07}            &{46.00 $\pm$ 32.52}                             & \textbf{74.85 $\pm$ 34.57}                                 &{38.25 $\pm$ 31.27}  \\
\bottomrule
\end{tabular}
}
\end{table*}

\begin{figure*}[!t]
    \centering
    \includegraphics[width=1.8\columnwidth]{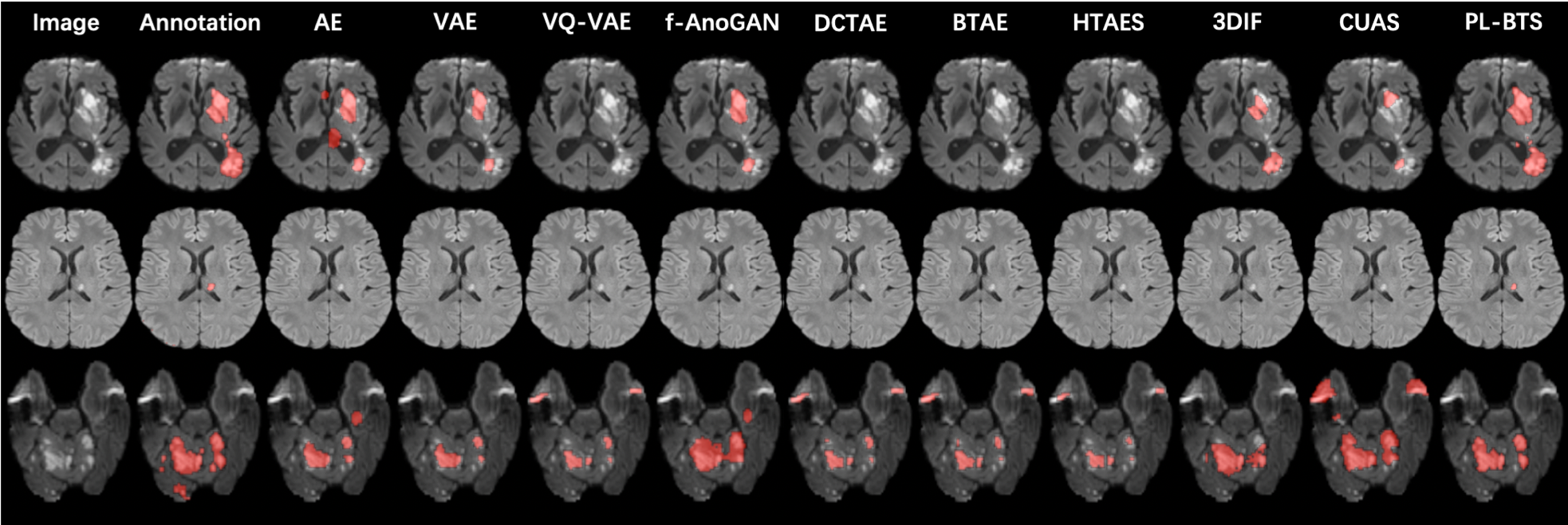}
    \caption{Examples of ischemic stroke lesion segmentation results (red) on test scans for the ISLES2022 dataset. The results of PL-BTS and each competing method are overlaid on the DWI. The expert annotation is also shown for reference.}
    \label{fig:ISLES}
\end{figure*}

\subsection{Comparison with competing unsupervised methods}

For fair comparison, since the competing unsupervised methods only use images without brain tumors or ischemic stroke lesions for training, their performance was compared with that of the prompt model (i.e., PL-BTS).

\subsubsection{Evaluation of brain tumor segmentation}

First, for qualitative evaluation, representative segmentation examples selected from the test sets of BraTS2021 and BTFLAIR are shown in Figs.~\ref{fig:BraTS021&BTFLAIR}a and \ref{fig:BraTS021&BTFLAIR}b, respectively, where the results of PL-BTS and the competing unsupervised methods are compared.
The manual annotation is also displayed for reference.
For both datasets our method produced tumor segmentation that is consistent with the annotation and better resembles the annotation than the results of the competing methods.

To quantitatively compare PL-BTS and each competing method, we computed the Dice coefficient, voxelwise precision, and voxelwise recall of the segmentation results on the test set for BraTS2021 and BTFLAIR.
The means and standard deviations of these evaluation metrics are listed in Table~\ref{tab:OASIS} for the two datasets separately, where paired Student's $t$-tests were also used to compare the results of PL-BTS and each competing method.
For both datasets and all three metrics, the result of PL-BTS is highly significantly ($p<0.001$) better than those of the competing methods. 
Compared with the second best method CUAS, PL-BTS has improved the Dice coefficient by about 45.3\% for BraTS2021 and 59.8\% for BTFLAIR.

In addition, to show the necessity of the validation task in PL-BTS, the quantitative results of PL-BTS achieved without the validation task (referred to as PL-BTS w/o VT) are shown in Table~\ref{tab:OASIS}. Compared with PL-BTS, the Dice coefficients of PL-BTS w/o VT drop by a large amount, which demonstrates the importance of using the validation task.

\subsubsection{Evaluation of ischemic stroke lesion segmentation}

For qualitative evaluation, representative segmentation cases for the test set of ISLES2022 are shown in Fig.~\ref{fig:ISLES} together with the manual annotation, where the results of PL-BTS and the competing methods are compared. The results of PL-BTS in general agree with the annotation and better resemble the annotation than those of the competing methods.

For quantitative evaluation, the means and standard deviations of the Dice coefficient, voxelwise precision, and voxelwise recall of the segmentation results on the test set are shown in Table~\ref{tab:ISLES}, where PL-BTS was compared with all competing methods with paired Student's $t$-tests.
The performance of PL-BTS is highly significantly ($p<0.001$) better than those of the competing methods in all cases. 
Compared with the second best method CUAS, PL-BTS has improved the Dice coefficient by about 141.4\%.

\begin{table}[!t]
\centering
\caption{Means and standard deviations of the Dice coefficients (\%), voxelwise precision (\%), and voxelwise recall (\%) of the segmentation results  of PL-BTS and the competing methods on the test set for ISLES2022. Asterisks indicate that the difference between PL-BTS and the competing method is statistically significant (***: $ p\le 0.001$) using a paired Student's $t$-test. The best results are highlighted in bold.}
\label{tab:ISLES}
\resizebox{1.0\columnwidth}{!}{
\begin{tabular}{cccc}
\toprule
\multirow{2}{*}{Method} & \multicolumn{3}{c}{\textbf{ISLES2022 (247 cases)}} \\\cmidrule(lr){2-4}
                         & Dice      & Precision    & Recall    \\
\midrule
\multirow{1}{*}{AE}      & 18.00 $\pm$ 21.45*** & 38.06 $\pm$ 39.93*** & 15.06 $\pm$ 15.06*** \\
\multirow{1}{*}{VAE}     & 13.10 $\pm$ 19.39*** & 45.84 $\pm$ 48.38***  & 8.499 $\pm$ 13.54***\\
\multirow{1}{*}{VQ-VAE} & 16.82 $\pm$ 23.49***   & 31.59 $\pm$ 36.23*** & 14.18 $\pm$ 20.95***\\
\multirow{1}{*}{f-AnoGAN} & 17.18 $\pm$ 20.23*** & 28.54 $\pm$ 35.39*** & 18.38 $\pm$ 21.16***\\
\multirow{1}{*}{DCTAE}  & 16.98 $\pm$ 23.48***  &30.29 $\pm$ 34.96***  & 14.65 $\pm$ 21.36***\\
\multirow{1}{*}{BTAE}   & 18.72 $\pm$ 25.02***  &  31.07 $\pm$ 35.14*** & 16.73 $\pm$ 23.41*** \\
\multirow{1}{*}{HTAES}  &14.26 $\pm$ 22.66***  & 32.20 $\pm$ 38.19*** & 11.35 $\pm$ 19.63*** \\
\multirow{1}{*}{3DIF}  & 14.28 $\pm$ 19.79***  & 17.37 $\pm$ 29.74*** & 35.56 $\pm$ 27.97***\\
\multirow{1}{*}{CUAS}   &21.31 $\pm$ 22.66***   & 21.11 $\pm$ 25.12*** & 42.20 $\pm$ 34.71*** \\
PL-BTS & \textbf{51.44 $\pm$ 34.58}  & \textbf{69.25 $\pm$ 32.16} & \textbf{47.06 $\pm$ 34.25}        
\\
\bottomrule
\end{tabular}
}
\end{table}

\subsection{Benefit of incorporating unlabeled images}
\label{sec:incorp_unlabeled}

We then used BraTS2021 and ISLES2022 to demonstrate the benefit of incorporating unlabeled images into model training.
Here, besides PL-BTS, the result of PL-BTS+ was considered. 
Also, to demonstrate the benefit of image pasting in PL-BTS+ in Section~\ref{sec:unlabeled}, the segmentation result achieved without the synthetic images obtained via pasting was considered, and it is referred to as PL-BTS+ without pasting undersegmented tumors~(PL-BTS+ w/o PUT).
In addition, the segmentation result achieved with the tumor pasting but without the original unlabeled images and pseudo-labels (referred to as PL-BTS+ w/o UIP) was evaluated to demonstrate the necessity of using the original unlabeled images. 

The Dice coefficients of PL-BTS and PL-BTS+ are summarized in Table~\ref{tab:ablation}, together with the results of PL-BTS+ w/o PUT and PL-BTS+ w/o UIP.
For both BraTS2021 and ISLES2022, PL-BTS+ outperforms PL-BTS, and 
it is also better than PL-BTS+ w/o PUT and PL-BTS+ w/o UIP.
These results indicate the benefit of incorporating unlabeled images for model training, as well as the benefit of using both original unlabeled images and images with pasted pseudo-labels.
Note that the Dice coefficient of PL-BTS+ w/o PUT for ISLES2022 is low and worse than PL-BTS.
This is likely because ischemic stroke lesions can be diffuse, and the false negatives in the pseudo-labels of the original unlabeled images can be severe and have a negative impact on model training. 

\begin{table}[!t]
\centering
\caption{Means and standard deviations of the Dice coefficients (\%) of the segmentation results of the proposed method on BraTS2021 (1151 cases) and ISLES2022 (147 cases) when unlabeled images were available for model training. The best results are highlighted in bold.}
\label{tab:ablation}
\resizebox{0.68\columnwidth}{!}{
\begin{tabular}{ccl}
\toprule
Dataset
& \multirow{-1}{*}{Method} & \multirow{-1}{*}{\quad \quad Dice} \\ \midrule

    \multirow{4}*{{BraTS2021}}
& PL-BTS           & 79.56 $\pm$ 15.01            \\
&PL-BTS+ w/o PUT   & 80.22 $\pm$ 15.07                \\
&PL-BTS+ w/o UIP   & 79.85 $\pm$ 16.22               \\                                                                                      

& PL-BTS+         & \textbf{82.66 $\pm$ 14.90}             \\ 
\midrule
\multirow{4}*{{ISLES2022}}
& PL-BTS       & 51.41 $\pm$ 33.88       \\
& PL-BTS+ w/o PUT   & 22.66 $\pm$ 28.80     \\
&PL-BTS+ w/o UIP   & 54.21 $\pm$ 33.39       \\                                                                                   
& PL-BTS+       &\textbf{55.29 $\pm$ 34.79}  \\ \bottomrule
\end{tabular}

}
\end{table}

\section{Discussion}
\label{sec:Dicscussion}
Our method PL-BTS for unsupervised brain tumor segmentation is based on image-based prompts that suggest whether tumor-like abnormality exists at a voxel.
It is different from existing reconstruction-based unsupervised approaches, where we artificially generate abnormality for model training.
Of note, the model training is nontrivial as the prompt model can easily overfit the simplistic synthetic training data.
To address this problem, we propose the use of a validation task, because no annotated data is available as a validation set. 
The results of brain tumor segmentation show that the new perspective presented in this work outperforms the existing methods by a large margin.
The use of the validation task may also inspire the development of zero-shot learning~\cite{xian2018zero} in general.
Model selection is an important issue in zero-shot learning, and the success of this work suggests that it is possible to use different tasks instead of annotated data to monitor the learning process, where the validation task should share general knowledge with the target task of interest.

The design of the prompt and validation tasks in this work is hand-crafted and relatively straightforward.
It is possible to explore more advanced designs of the tasks.
For example, the diversity of the data generated for the prompt and validation tasks may be increased to improve the stability of model training and selection.
In addition, the use of multiple validation tasks may be explored in future works, where these tasks can synergistically avoid overfitting with proper information fusion.
Finally, it is an interesting direction to explore learning-based task design, where the generation of abnormality in both prompt and validation tasks can be parameterized and the parameters are learned jointly in the model training process.


As there is currently not yet a vision-language model for 3D brain MR images, our prompt design is purely image-based.
However, since these images are usually accompanied by radiological reports, it is possible to train a vision-language model for brain tumors with such data, and the model can then be used for prompt-based tumor segmentation.
The development of such models can be explored in future works. 

Although the proposed method uses a different strategy than existing reconstruction-based methods, it is possible to integrate these different strategies. 
For example, currently, the generation of abnormality is performed randomly in brain tissue.
The reconstruction-based approach may provide information about where it is easy or difficult to reconstruct normal tissue.
This can partition normal tissue into different groups, and the location for abnormality generation can be determined based on the discrimination.

Besides brain tumor segmentation, we have demonstrated the potential of extending the proposed method to other brain lesion segmentation tasks.
In particular, we applied our method to ischemic stroke lesion segmentation and obtained results that are similar to those of brain tumor segmentation.
Future works can explore additional brain lesion segmentation tasks and make customized designs for them.

In addition to PL-BTS, we have proposed PL-BTS+ that extends PL-BTS when unlabeled images with brain tumors are available. 
The extension is motivated by the pseudo-labeling strategy that is commonly used in semi-supervised learning, and based on our observation that the pseudo-labels given by the prompt model tend to undersegment tumors, we further paste the undersegmented tumors onto tumor-free images to reduce false negatives.
The results presented in Section~\ref{sec:incorp_unlabeled} confirm the benefit of the proposed extension, as well as the necessity of using both original unlabeled images and pasted images.
The use of unlabeled images also motivates us to further explore ``semi-unsupervised'' segmentation of brain tumors and other types of lesions in future works, where normal-appearing images can be used jointly with unlabeled images with lesions to improve model training.
Approaches that are commonly used in semi-supervised learning may be adapted to the semi-unsupervised setting.

\section{Conclusion}
\label{sec:conclusion}
We have proposed an unsupervised DL-based brain tumor segmentation method PL-BTS, which is inspired by prompt learning in NLP.
Image-based prompts are designed to train a prompt model that suggests tumor-like abnormality, where we also propose to use a validation task to avoid overfitting. 
In addition, we extend PL-BTS and propose PL-BTS+, which allows further improvement of the segmentation performance  when unlabeled images with brain tumors are available. 
Experiments on public and in-house datasets show that our method outperforms existing unsupervised DL-based methods by a large margin and it can potentially be extended to other brain lesion segmentation tasks as well.

\bibliographystyle{IEEEtran}
\bibliography{refs}

\begin{thebibliography}{10}
\providecommand{\url}[1]{#1}
\csname url@samestyle\endcsname
\providecommand{\newblock}{\relax}
\providecommand{\bibinfo}[2]{#2}
\providecommand{\BIBentrySTDinterwordspacing}{\spaceskip=0pt\relax}
\providecommand{\BIBentryALTinterwordstretchfactor}{4}
\providecommand{\BIBentryALTinterwordspacing}{\spaceskip=\fontdimen2\font plus
\BIBentryALTinterwordstretchfactor\fontdimen3\font minus
  \fontdimen4\font\relax}
\providecommand{\BIBforeignlanguage}[2]{{%
\expandafter\ifx\csname l@#1\endcsname\relax
\typeout{** WARNING: IEEEtran.bst: No hyphenation pattern has been}%
\typeout{** loaded for the language `#1'. Using the pattern for}%
\typeout{** the default language instead.}%
\else
\language=\csname l@#1\endcsname
\fi
#2}}
\providecommand{\BIBdecl}{\relax}
\BIBdecl

\bibitem{Andrea2022Spatiotemporal}
A.~Comba \emph{et~al.}, ``Spatiotemporal analysis of glioma heterogeneity
  reveals {COL1A1} as an actionable target to disrupt tumor progression,''
  \emph{Nature Communications}, vol.~13, no.~1, p. 3606, 2022.

\bibitem{Menze2015Multimodal}
B.~H. Menze \emph{et~al.}, ``The multimodal brain tumor image segmentation
  benchmark ({BRATS}),'' \emph{IEEE Transactions on Medical Imaging}, vol.~34,
  no.~10, pp. 1993--2024, 2015.

\bibitem{Williams2008Science}
D.~W. Parsons \emph{et~al.}, ``An integrated genomic analysis of human
  glioblastoma multiforme,'' \emph{Science}, vol. 321, no. 5897, pp.
  1807--1812, 2008.

\bibitem{Gregory2020Systemic}
J.~V. Gregory \emph{et~al.}, ``Systemic brain tumor delivery of synthetic
  protein nanoparticles for glioblastoma therapy,'' \emph{Nature
  Communications}, vol.~11, no.~1, p. 5687, 2020.

\bibitem{ALIFIERIS201563}
C.~Alifieris and D.~T. Trafalis, ``{Glioblastoma multiforme: Pathogenesis and
  treatment},'' \emph{Pharmacology and Therapeutics}, vol. 152, pp. 63--82,
  2015.

\bibitem{isensee2021nnu}
F.~Isensee and K.~H. Maier-Hein, ``{nnU-Net} for brain tumor segmentation,'' in
  \emph{MICCAI BrainLes Workshop}, vol. 12659.\hskip 1em plus 0.5em minus
  0.4em\relax Springer, 2020, p. 118.

\bibitem{magadza2021deep}
T.~Magadza and S.~Viriri, ``Deep learning for brain tumor segmentation: a
  survey of state-of-the-art,'' \emph{Journal of Imaging}, vol.~7, no.~2,
  p.~19, 2021.

\bibitem{kamnitsas2017efficient}
K.~Kamnitsas \emph{et~al.}, ``Efficient multi-scale {3D CNN} with fully
  connected {CRF} for accurate brain lesion segmentation,'' \emph{Medical Image
  Analysis}, vol.~36, pp. 61--78, 2017.

\bibitem{gordillo2013state}
N.~Gordillo, E.~Montseny, and P.~Sobrevilla, ``State of the art survey on {MRI}
  brain tumor segmentation,'' \emph{Magnetic Resonance Imaging}, vol.~31,
  no.~8, pp. 1426--1438, 2013.

\bibitem{dubey2009region}
R.~Dubey, M.~Hanmandlu, S.~Gupta, and S.~Gupta, ``Region growing for {MRI}
  brain tumor volume analysis,'' \emph{Indian Journal of Science and
  Technology}, vol.~2, no.~9, 2009.

\bibitem{havaei2017brain}
M.~Havaei \emph{et~al.}, ``Brain tumor segmentation with deep neural
  networks,'' \emph{Medical Image Analysis}, vol.~35, pp. 18--31, 2017.

\bibitem{wang2021transbts}
W.~Wang, C.~Chen, M.~Ding, H.~Yu, S.~Zha, and J.~Li, ``{TransBTS}: Multimodal
  brain tumor segmentation using transformer,'' in \emph{International
  Conference on Medical Image Computing and Computer-Assisted
  Intervention}.\hskip 1em plus 0.5em minus 0.4em\relax Springer, 2021, pp.
  109--119.

\bibitem{ding2021rfnet}
Y.~Ding, X.~Yu, and Y.~Yang, ``{RFNet}: Region-aware fusion network for
  incomplete multi-modal brain tumor segmentation,'' in \emph{Proceedings of
  the IEEE/CVF International Conference on Computer Vision}, 2021, pp.
  3975--3984.

\bibitem{ding2021tostagan}
Y.~Ding \emph{et~al.}, ``{ToStaGAN}: An end-to-end two-stage generative
  adversarial network for brain tumor segmentation,'' \emph{Neurocomputing},
  vol. 462, pp. 141--153, 2021.

\bibitem{cui2019semi}
W.~Cui \emph{et~al.}, ``Semi-supervised brain lesion segmentation with an
  adapted mean teacher model,'' in \emph{International Conference on
  Information Processing in Medical Imaging}.\hskip 1em plus 0.5em minus
  0.4em\relax Springer, 2019, pp. 554--565.

\bibitem{meier2014patient}
R.~Meier, S.~Bauer, J.~Slotboom, R.~Wiest, and M.~Reyes, ``Patient-specific
  semi-supervised learning for postoperative brain tumor segmentation,'' in
  \emph{International Conference on Medical Image Computing and
  Computer-Assisted Intervention}.\hskip 1em plus 0.5em minus 0.4em\relax
  Springer, 2014, pp. 714--721.

\bibitem{zhang2021self}
X.~Zhang, W.~Xie, C.~Huang, Y.~Zhang, and Y.~Wang, ``Self-supervised tumor
  segmentation through layer decomposition,'' \emph{arXiv preprint
  arXiv:2109.03230}, 2021.

\bibitem{zhuang2019self}
X.~Zhuang, Y.~Li, Y.~Hu, K.~Ma, Y.~Yang, and Y.~Zheng, ``Self-supervised
  feature learning for {3D} medical images by playing a rubik's cube,'' in
  \emph{International Conference on Medical Image Computing and
  Computer-Assisted Intervention}.\hskip 1em plus 0.5em minus 0.4em\relax
  Springer, 2019, pp. 420--428.

\bibitem{pinaya2022unsupervised}
W.~H. Pinaya \emph{et~al.}, ``Unsupervised brain imaging {3D} anomaly detection
  and segmentation with transformers,'' \emph{Medical Image Analysis}, vol.~79,
  p. 102475, 2022.

\bibitem{baur2021autoencoders}
C.~Baur, S.~Denner, B.~Wiestler, N.~Navab, and S.~Albarqouni, ``Autoencoders
  for unsupervised anomaly segmentation in brain {MR} images: a comparative
  study,'' \emph{Medical Image Analysis}, vol.~69, p. 101952, 2021.

\bibitem{naval2021implicit}
S.~Naval~Marimont and G.~Tarroni, ``Implicit field learning for unsupervised
  anomaly detection in medical images,'' in \emph{International Conference on
  Medical Image Computing and Computer-Assisted Intervention}.\hskip 1em plus
  0.5em minus 0.4em\relax Springer, 2021, pp. 189--198.

\bibitem{wu2021unsupervised}
X.~Wu, L.~Bi, M.~Fulham, D.~D. Feng, L.~Zhou, and J.~Kim, ``Unsupervised brain
  tumor segmentation using a symmetric-driven adversarial network,''
  \emph{Neurocomputing}, vol. 455, pp. 242--254, 2021.

\bibitem{liu2021pre}
P.~Liu, W.~Yuan, J.~Fu, Z.~Jiang, H.~Hayashi, and G.~Neubig, ``Pre-train,
  prompt, and predict: A systematic survey of prompting methods in natural
  language processing,'' \emph{arXiv preprint arXiv:2107.13586}, 2021.

\bibitem{kenton2019bert}
J.~D. M.-W.~C. Kenton and L.~K. Toutanova, ``{BERT}: Pre-training of deep
  bidirectional transformers for language understanding,'' in \emph{Proceedings
  of NAACL-HLT}, 2019, pp. 4171--4186.

\bibitem{brown2020language}
T.~Brown \emph{et~al.}, ``Language models are few-shot learners,'' in
  \emph{Advances in Neural Information Processing Systems}, vol.~33, 2020, pp.
  1877--1901.

\bibitem{bahng2022visual}
H.~Bahng, A.~Jahanian, S.~Sankaranarayanan, and P.~Isola, ``{Visual Prompting}:
  Modifying pixel space to adapt pre-trained models,'' \emph{arXiv preprint
  arXiv:2203.17274}, 2022.

\bibitem{changpinyo2017predicting}
S.~Changpinyo, W.-L. Chao, and F.~Sha, ``Predicting visual exemplars of unseen
  classes for zero-shot learning,'' in \emph{Proceedings of the IEEE
  International Conference on Computer Vision}, 2017, pp. 3476--3485.

\bibitem{mutasa2020understanding}
S.~Mutasa, S.~Sun, and R.~Ha, ``Understanding artificial intelligence based
  radiology studies: What is overfitting?'' \emph{Clinical Imaging}, vol.~65,
  pp. 96--99, 2020.

\bibitem{ghiasi2021simple}
G.~Ghiasi \emph{et~al.}, ``Simple copy-paste is a strong data augmentation
  method for instance segmentation,'' in \emph{IEEE/CVF Conference on Computer
  Vision and Pattern Recognition}.\hskip 1em plus 0.5em minus 0.4em\relax IEEE,
  2021, pp. 2917--2927.

\bibitem{li2020analyzing}
Z.~Li, K.~Kamnitsas, and B.~Glocker, ``Analyzing overfitting under class
  imbalance in neural networks for image segmentation,'' \emph{IEEE
  Transactions on Medical Imaging}, vol.~40, no.~3, pp. 1065--1077, 2020.

\bibitem{dong2017automatic}
H.~Dong, G.~Yang, F.~Liu, Y.~Mo, and Y.~Guo, ``Automatic brain tumor detection
  and segmentation using {U-Net} based fully convolutional networks,'' in
  \emph{Annual Conference on Medical Image Understanding and Analysis}.\hskip
  1em plus 0.5em minus 0.4em\relax Springer, 2017, pp. 506--517.

\bibitem{hu2022mutual}
J.~Hu, X.~Gu, and X.~Gu, ``Mutual ensemble learning for brain tumor
  segmentation,'' \emph{Neurocomputing}, vol. 504, pp. 68--81, 2022.

\bibitem{supot2007segmentation}
S.~Supot, C.~Thanapong, P.~Chuchart, and S.~Manas, ``Segmentation of magnetic
  resonance images using discrete curve evolution and fuzzy clustering,'' in
  \emph{IEEE International Conference on Integration Technology}.\hskip 1em
  plus 0.5em minus 0.4em\relax IEEE, 2007, pp. 697--700.

\bibitem{abdel2015brain}
E.~Abdel-Maksoud, M.~Elmogy, and R.~Al-Awadi, ``Brain tumor segmentation based
  on a hybrid clustering technique,'' \emph{Egyptian Informatics Journal},
  vol.~16, no.~1, pp. 71--81, 2015.

\bibitem{jiang2022swinbts}
Y.~Jiang, Y.~Zhang, X.~Lin, J.~Dong, T.~Cheng, and J.~Liang, ``{SwinBTS}: a
  method for {3D} multimodal brain tumor segmentation using {Swin
  Transformer},'' \emph{Brain Sciences}, vol.~12, no.~6, p. 797, 2022.

\bibitem{hatamizadeh2022unetr}
A.~Hatamizadeh \emph{et~al.}, ``{UNETR}: Transformers for {3D} medical image
  segmentation,'' in \emph{Proceedings of the IEEE/CVF Winter Conference on
  Applications of Computer Vision}, 2022, pp. 574--584.

\bibitem{3dunet}
{\"O}.~{\c{C}}i{\c{c}}ek, A.~Abdulkadir, S.~S. Lienkamp, T.~Brox, and
  O.~Ronneberger, ``{3D} {U-Net}: Learning dense volumetric segmentation from
  sparse annotation,'' in \emph{International Conference on Medical Image
  Computing and Computer-Assisted Intervention}.\hskip 1em plus 0.5em minus
  0.4em\relax Springer, 2016, pp. 424--432.

\bibitem{2dunet}
O.~Ronneberger, P.~Fischer, and T.~Brox, ``U-{N}et: Convolutional networks for
  biomedical image segmentation,'' in \emph{{I}nternational {C}onference on
  {M}edical {I}mage {C}omputing and {C}omputer-{A}ssisted
  {I}ntervention}.\hskip 1em plus 0.5em minus 0.4em\relax Springer, 2015, pp.
  234--241.

\bibitem{luo2021dtc}
X.~Luo, J.~Chen, T.~Song, and G.~Wang, ``Semi-supervised medical image
  segmentation through dual-task consistency,'' in \emph{AAAI Conference on
  Artificial Intelligence}, 2021, pp. 8801--8809.

\bibitem{chen2019self}
L.~Chen, P.~Bentley, K.~Mori, K.~Misawa, M.~Fujiwara, and D.~Rueckert,
  ``Self-supervised learning for medical image analysis using image context
  restoration,'' \emph{Medical Image Analysis}, vol.~58, p. 101539, 2019.

\bibitem{dey2021asc}
R.~Dey and Y.~Hong, ``{ASC-Net}: Adversarial-based selective network for
  unsupervised anomaly segmentation,'' in \emph{International Conference on
  Medical Image Computing and Computer-Assisted Intervention}.\hskip 1em plus
  0.5em minus 0.4em\relax Springer, 2021, pp. 236--247.

\bibitem{marimont2021anomaly}
S.~N. Marimont and G.~Tarroni, ``Anomaly detection through latent space
  restoration using vector quantized variational autoencoders,'' in
  \emph{International Symposium on Biomedical Imaging}.\hskip 1em plus 0.5em
  minus 0.4em\relax IEEE, 2021, pp. 1764--1767.

\bibitem{wang2020image}
L.~Wang, D.~Zhang, J.~Guo, and Y.~Han, ``Image anomaly detection using normal
  data only by latent space resampling,'' \emph{Applied Sciences}, vol.~10,
  no.~23, p. 8660, 2020.

\bibitem{schlegl2019fAnogan}
T.~Schlegl, P.~Seeb{\"o}ck, S.~M. Waldstein, G.~Langs, and U.~Schmidt-Erfurth,
  ``{f-AnoGAN}: Fast unsupervised anomaly detection with generative adversarial
  networks,'' \emph{Medical Image Analysis}, vol.~54, pp. 30--44, 2019.

\bibitem{git2022Ahmed}
A.~Ghorbel, A.~Aldahdooh, S.~Albarqouni, and W.~Hamidouche, ``Transformer based
  models for unsupervised anomaly segmentation in brain {MR} images,''
  \emph{arXiv preprint arXiv:2207.02059}, 2022.

\bibitem{xu2021simple}
M.~Xu \emph{et~al.}, ``A simple baseline for zero-shot semantic segmentation
  with pre-trained vision-language model,'' \emph{arXiv preprint
  arXiv:2112.14757}, 2021.

\bibitem{ding2022decoupling}
J.~Ding, N.~Xue, G.-S. Xia, and D.~Dai, ``Decoupling zero-shot semantic
  segmentation,'' in \emph{Proceedings of the IEEE/CVF Conference on Computer
  Vision and Pattern Recognition}, 2022, pp. 11\,583--11\,592.

\bibitem{wang2022cris}
Z.~Wang \emph{et~al.}, ``{CRIS}: {CLIP}-driven referring image segmentation,''
  in \emph{Proceedings of the IEEE/CVF Conference on Computer Vision and
  Pattern Recognition}, 2022, pp. 11\,686--11\,695.

\bibitem{radford2021CLIP}
A.~Radford \emph{et~al.}, ``Learning transferable visual models from natural
  language supervision,'' in \emph{International Conference on Machine
  Learning}.\hskip 1em plus 0.5em minus 0.4em\relax PMLR, 2021, pp. 8748--8763.

\bibitem{perez2021self}
F.~Pérez-García~F \emph{et~al.}, ``A self-supervised learning strategy for
  postoperative brain cavity segmentation simulating resections,''
  \emph{International Journal of Computer Assisted Radiology and Surgery},
  vol.~16, no.~10, pp. 1653--1661, 2021.

\bibitem{Diederik2015Adam}
D.~P. Kingma and J.~Ba, ``Adam: A method for stochastic optimization,'' in
  \emph{International Conference on Learning Representations}, 2015.

\bibitem{zoph2020rethinking}
B.~Zoph \emph{et~al.}, ``Rethinking pre-training and self-training,''
  \emph{Advances in Neural Information Processing Systems}, vol.~33, pp.
  3833--3845, 2020.

\bibitem{marcus2010open}
D.~S. Marcus, A.~F. Fotenos, J.~G. Csernansky, J.~C. Morris, and R.~L. Buckner,
  ``Open access series of imaging studies: longitudinal {MRI} data in
  nondemented and demented older adults,'' \emph{Journal of Cognitive
  Neuroscience}, vol.~22, no.~12, pp. 2677--2684, 2010.

\bibitem{baid2021rsna}
U.~Baid \emph{et~al.}, ``The {RSNA-ASNR-MICCAI BraTS} 2021 benchmark on brain
  tumor segmentation and radiogenomic classification,'' \emph{arXiv preprint
  arXiv:2107.02314}, 2021.

\bibitem{ISLES}
M.~R.~H. Petzsche \emph{et~al.}, ``{ISLES} 2022: A multi-center magnetic
  resonance imaging stroke lesion segmentation dataset,'' \emph{arXiv preprint
  arXiv:2206.06694}, 2022.

\bibitem{smith2002fast}
S.~M. Smith, ``Fast robust automated brain extraction,'' \emph{Human brain
  mapping}, vol.~17, no.~3, pp. 143--155, 2002.

\bibitem{hakimelahi2014time}
R.~Hakimelahi \emph{et~al.}, ``Time and diffusion lesion size in major anterior
  circulation ischemic strokes,'' \emph{Stroke}, vol.~45, no.~10, pp.
  2936--2941, 2014.

\bibitem{silva2022constrained}
J.~Silva-Rodr{\'\i}guez, V.~Naranjo, and J.~Dolz, ``Constrained unsupervised
  anomaly segmentation,'' \emph{arXiv preprint arXiv:2203.01671}, 2022.

\bibitem{xian2018zero}
Y.~Xian, C.~H. Lampert, B.~Schiele, and Z.~Akata, ``Zero-shot learning a
  comprehensive evaluation of the good, the bad and the ugly,'' \emph{IEEE
  transactions on pattern analysis and machine intelligence}, vol.~41, no.~9,
  pp. 2251--2265, 2018.

\end{thebibliography}

\end{document}